\documentclass[letterpaper, 10 pt, conference]{ieeeconf}  
\usepackage{graphicx}    
\usepackage{array}       
\usepackage{booktabs}    
\usepackage{tabularx}    
\usepackage{rotating}    
\usepackage{multirow}
\usepackage{cellspace}
 \usepackage{amsmath} 
\usepackage{float}       
\usepackage{cite}
\usepackage[table]{xcolor}
\newcommand{\sqrgb}[3]{\textcolor[rgb]{#1,#2,#3}{\rule{1.2ex}{1.2ex}}}

\IEEEoverridecommandlockouts                              
\usepackage[utf8]{inputenc}
\usepackage[T1]{fontenc}
\usepackage{amssymb} 

\makeatletter
\let\NAT@parse\undefined
\makeatother
\usepackage[pagebackref=false,breaklinks=true,colorlinks,bookmarks=false]{hyperref}

\def\ie{\emph{i.e.}}

\title{\LARGE \bf
Exploring Single Domain Generalization of LiDAR-based Semantic Segmentation under Imperfect Labels
}

\author{Weitong Kong$^{1,*}$, Zichao Zeng$^{2,*}$, Di Wen$^{1}$, Jiale Wei$^{1}$, \\
Kunyu Peng$^{1,\dag}$, June Moh Goo$^{2}$, Jan Boehm$^{2}$, and Rainer Stiefelhagen$^{1}$
\thanks{*\;Equal contribution}%
\thanks{\dag\;Corresponding Author (email:\href{mailto:kunyu.peng@kit.edu}{kunyu.peng@kit.edu})}%
\thanks{\raggedright{$^{1}$ The authors are with Institute for Anthropomatics and Robotics, Karlsruhe Institute of Technology, Karlsruhe 76131, Germany.}}
\thanks{$^{2}$ The authors are with Department of Civil, Environmental and Geomatic Engineering, University College London, London  WC1E 6BT, U.K.}
\thanks{The project is funded by the Deutsche Forschungsgemeinschaft (DFG, German Research Foundation) – SFB-1574 – 471687386. This work was supported in part by the SmartAge project sponsored by the Carl Zeiss Stiftung (P2019-01-003; 2021-2026). The authors gratefully acknowledge the computing time provided on the high-performance computer HoreKa by the National High-Performance Computing Center at KIT (NHR@KIT). This center is jointly supported by the Federal Ministry of Education and Research and the Ministry of Science, Research and the Arts of Baden-Württemberg, as part of the National High-Performance Computing (NHR) joint funding program (https://www.nhr-verein.de/en/our-partners). HoreKa is partly funded by the German Research Foundation (DFG).
}
}

\begin{document}

\maketitle
\thispagestyle{empty}
\pagestyle{empty}

\begin{abstract}
Accurate perception is critical for vehicle safety, with LiDAR as a key enabler in autonomous driving. To ensure robust performance across environments, sensor types, and weather conditions without costly re-annotation, domain generalization in LiDAR-based 3D semantic segmentation is essential.
However, LiDAR annotations are often noisy due to sensor imperfections, occlusions, and human errors. Such noise degrades segmentation accuracy and is further amplified under domain shifts, threatening system reliability. While noisy-label learning is well-studied in images, its extension to 3D LiDAR segmentation under domain generalization remains largely unexplored, as the sparse and irregular structure of point clouds limits direct use of 2D methods.
To address this gap, we introduce the novel task Domain Generalization for LiDAR Semantic Segmentation under Noisy Labels (DGLSS-NL) and establish the first benchmark by adapting three representative noisy-label learning strategies from image classification to 3D segmentation.
However, we find that existing noisy-label learning approaches adapt poorly to LiDAR data. We therefore propose DuNe, a dual-view framework with strong and weak branches that enforce feature-level consistency and apply cross-entropy loss based on confidence-aware filtering of predictions.
Our approach shows state-of-the-art performance by achieving \textbf{56.86\%} mIoU on SemanticKITTI, \textbf{42.28\%} on nuScenes, and \textbf{52.58\%} on SemanticPOSS under 10\% symmetric label noise, with an overall Arithmetic Mean (AM) of \textbf{49.57\%} and Harmonic Mean (HM) of \textbf{48.50\%}, thereby demonstrating robust domain generalization in DGLSS-NL tasks. The code is available at \href{https://github.com/MKong17/DGLSS-NL.git}{https://github.com/MKong17/DGLSS-NL.git}.
\end{abstract}
\section{INTRODUCTION}
In autonomous driving, system safety and reliability are paramount. Accurate perception is essential to this goal. LiDAR provides precise 3D geometry and is central to reliable perception~\cite{bib:lidar-survey, peng2022mass}. However, the quality of LiDAR data varies across sensors, environments, and operating conditions, which challenges consistent interpretation. Domain generalization in 3D LiDAR semantic segmentation trains models that must perform in unseen domains without target data~\cite{kim2023single}. This capability is critical for robust perception and safe decision making in diverse real-world scenarios~\cite{fu2023styleadv,fu2024cross}.

Most existing domain generalization methods assume perfect annotations. In practice, 3D LiDAR labels are often imperfect because dense and irregular point sets are difficult to annotate consistently~\cite{behley2019semantickitti,bib:waymo,LabelImbalance}. Label noise degrades segmentation performance, and the degradation is amplified under domain shift~\cite{bib:label-noise-survey}, which threatens reliability once deployed. These observations motivate a setting that treats noisy supervision and domain shift jointly.

\begin{figure}
    \includegraphics[width=1\linewidth]{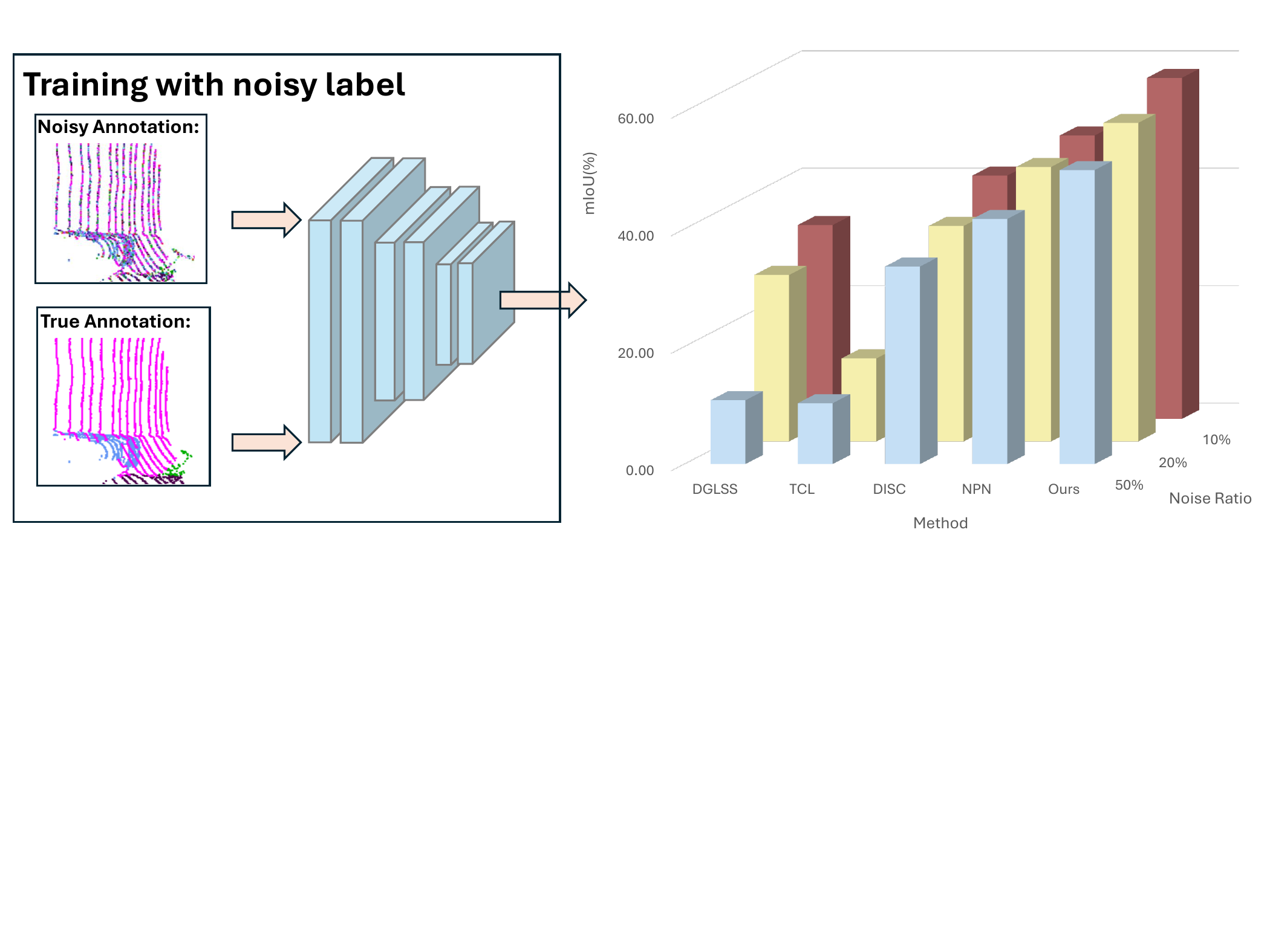}
    \vspace{-100pt}
    \caption{
    We inject symmetric label noise into the training set according to predefined noise ratios. 
The figure reports the segmentation performance on the test set, where the vertical axis shows mIoU(\%)($\uparrow$), 
the horizontal axis corresponds to the applied methods, 
and the third dimension indicates the results under different ratios of symmetric noise. 
\textbf{DuNe} consistently achieves the best results across all noise levels.
    }
    \vskip-3ex
    \label{fig:placeholder}
\end{figure}

In the image domain, noisy-label learning has been extensively studied, including loss correction, sample reweighting, and semi-supervised or contrastive strategies. 
TCL~\cite{bib:tcl}, a representative method, applies Gaussian modeling and pseudo-labeling to enable contrastive learning.
On the other hand, DISC~\cite{bib:DISC} dynamically selects clean samples while simultaneously exploiting informative signals from noisy ones. 
NPN~\cite{bib:npn} accumulates prediction statistics to construct partial and negative label sets, thereby mitigating the effect of overconfident noise.
Although these approaches demonstrate strong performance on images, directly transferring them to point clouds is non-trivial. Unlike images, point clouds are sparse, irregular, and orderless~\cite{bib:PointNet,wu2024rethinking}, which necessitates adaptations that respect the unique characteristics of 3D geometry.

This work presents the first systematic study of domain generalization for LiDAR semantic segmentation under noisy labels. We construct a controlled benchmark by corrupting source-domain labels with symmetric noise at predefined ratios, and evaluate both in-domain and cross-domain without accessing target samples or statistics. To ground the study, we adapt three representative noisy-label strategies from images to large-scale point clouds, namely TCL~\cite{bib:tcl}, DISC~\cite{bib:DISC}, and NPN~\cite{bib:npn}, on a unified LiDAR backbone with matched optimization and augmentation. Building on empirical findings, we introduce \textbf{DuNe}, a \textbf{Du}al-view framework for learning with \textbf{N}oisy lab\textbf{e}ls in 3D point clouds. It couples a strong geometry-aware view with a weak view, aligns them via bottleneck consistency, and employs confidence-filtered partial and negative supervision.

\begin{itemize}
    \item \textbf{DGLSS-NL benchmark}. We establish the DGLSS-NL benchmark for single-source LiDAR semantic segmentation under controlled symmetric label noise, with rigorous in-domain and cross-domain evaluation to enable reproducible and fair comparison.
    \item \textbf{Standardized transfers and diagnostic insights}. We adapt three representative noisy-label strategies from images (TCL, DISC, NPN) to large-scale point clouds on a unified LiDAR backbone with matched training recipes, and conduct a controlled diagnostic analysis that disentangles the roles of sample selection, contrastive objectives, and negative learning under domain shift and varying noise severities, yielding actionable insights for future 3D methods.
    \item \textbf{Dual-view noise-robust generalization}. We introduce \textbf{DuNe}, a dual-view framework that fuses a geometry-aware strong view with a complementary weak view and integrates noise-aware supervision to resist label corruption and shift. The resulting model consistently surpasses the strongest transferred baseline across datasets and noise levels, improving Arithmetic Mean (AM) and Harmonic Mean (HM) of overall mIoU percentages by +4.6 / +4.6 at 10\% noise, +6.2 / +6.9 at 20\% noise, and +8.6 / +8.2 at 50\% noise on average. These results are obtained by training on SemanticKITTI and evaluating on SemanticKITTI, nuScenes, and SemanticPOSS (denoted as K$\rightarrow$K/N/P).
\end{itemize}

\section{Related Work}
\subsection{LiDAR Semantic Segmentation} 
LiDAR semantic segmentation (LSS) assigns semantic categories to 3D points, with existing methods being point-~\cite{bib:PointNet, fei2021pillarsegnet}, projection-~\cite{wu2018squeezeseg}, or voxel-~\cite{zhou2018voxelnet} based.
Point-based methods directly process raw points and capture fine-grained local geometry but scale poorly to large outdoor scenes~\cite{bib:PointNet,qi2017pointnetplusplus}. 
Projection-based methods transform point clouds into 2D views, enabling efficient processing with standard CNNs and even off-the-shelf vision transformers, though inevitably at the cost of geometric distortion and information loss\cite{wu2018squeezeseg,milioto2019rangenet++,June3DSeg}.
Voxel-based methods discretize the 3D space and apply sparse convolutions to balance accuracy and efficiency~\cite{zhou2018voxelnet,graham20183d}. 
Hybrid designs combine multiple representations for improved robustness~\cite{hu2020randla}. 
Despite these advances, current models still rely heavily on costly manual annotations~\cite{behley2019semantickitti,caesar2020nuscenes,pan2020poss}, 
which are often imperfect, imbalance, prone to annotation noise, and amplify sensitivity to dataset bias \cite{LabelImbalance}.
\subsection{Domain Generalization in LiDAR Perception}
While LSS models have achieved notable performance, they often degrade severely when applied to unseen environments, highlighting the need for 3D domain generalization (3DDG).
Early studies such as MetaSets~\cite{huang2021metasets} addressed 3DDG for classification, where meta-learning with geometry-based transformations was introduced to bridge the gap between synthetic and real point clouds. Similarly, MAL~\cite{huang2022manifold} expanded the source domain with adversarially generated intermediate domains on transformation manifolds, improving generalization to unseen target sets.
Beyond classification, Domain Generalization for LiDAR Semantic Segmentation (DGLSS) module~\cite{kim2023single} formalized DG in large-scale outdoor LiDAR. DGLSS addressed sparsity variations across sensors and scene distribution shifts challenges and proposed sparsity augmentation together with consistency regularizations. This established the first benchmark for LiDAR DG.
However, existing DGLSS approaches assume clean supervision. In practice, large-scale LiDAR annotations are costly, imperfect, and inevitably noisy. This motivates our work on DGLSS under noisy labels (DGLSS-NL), where robustness to annotation noise is jointly addressed with domain generalization.

\subsection{Noisy-label Learning}
Although domain generalization alleviates dataset bias, its assumption of clean annotations rarely holds in real-world LiDAR datasets, where supervision is frequently corrupted by occlusion, sparsity, long-range effects, or human errors~\cite{behley2019semantickitti,caesar2020nuscenes,pan2020poss}. 
Noisy-label learning addresses this issue by enabling reliable training under imperfect annotations. 
In the image domain, research has focused on loss modification~\cite{patrini2017forward,jiang2018mentornet, bib:tcl}, regularization strategies~\cite{ghosh2017robust,ma2020normalized}, and dynamic sample selection~\cite{bib:sample-selection, bib:DISC, bib:npn}. 
Despite progress, transferring existing techniques to LiDAR segmentation is challenging due to irregular points, costly annotations, and geometric constraints. LSS faces noisy labels, overlooked by domain generalization methods, while noise-robust approaches rarely address domain shifts. This motivates our noise-robust domain generalization framework for large-scale 3D point clouds.

\begin{figure*}
    \centering
    \includegraphics[width=0.90\textwidth]{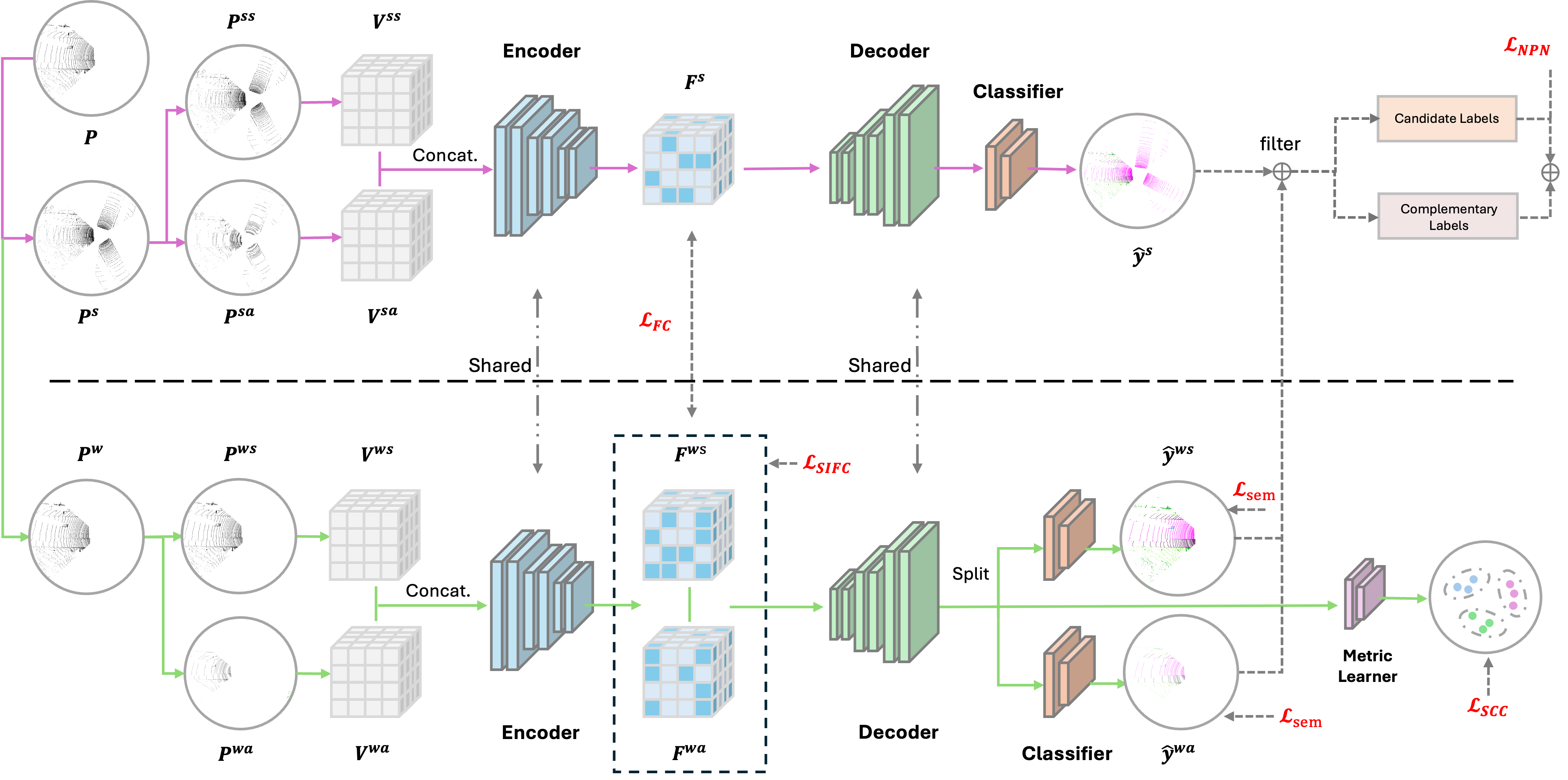}
    \caption{\textbf{Overview of our proposed dual-view training pipeline - DuNe.}
    Each input LiDAR scan is first augmented by PolarMix~\cite{xiao2022polarmix} to generate a 
    \textit{strong} view and a \textit{weak} view. 
    Both views are further processed with sparsity augmentation. 
    For the strong view, the entire augmented scan is used to construct candidate labels and complementary labels~\cite{bib:npn}, which provide noise-robust supervision. 
    For the weak view, we explicitly split the original and sparsity-augmented versions to form paired inputs, and enforce consistency loss and semantic correlation loss between them. 
    This design allows the framework to jointly exploit label-level robustness and view-level consistency for improved noisy-label learning.}
    \label{fig:placeholder2}
    \vskip-3ex
\end{figure*}
\section{Benchmark and Baselines}

\subsection{Noisy Labels}\label{sec:nl}
In our benchmark, we introduce noisy labels into existing domain generalization models for 3D point cloud semantic segmentation by adapting strategies from the image domain, 
aiming to evaluate model robustness under different noise levels. 
To simulate noise, we replace the ground-truth labels of certain samples with other categories in the dataset, following standard noisy-label protocols~\cite{patrini2017forward}.  

We distinguish between two types of noise: symmetric and asymmetric. 
Symmetric noise randomly flips each label to another class with equal probability, 
while asymmetric noise better reflects real-world annotation errors, where mislabeling is biased toward semantically similar categories (e.g., ``truck'' $\rightarrow$ ``bus''). 
Since asymmetric noise requires prior knowledge of class-level confusion statistics, we focus only on symmetric noise as a principled starting point. 
In practice, small noise (e.g., 2\% or 5\%) levels cause little disturbance due to the inherent robustness of the backbone, 
while very high noise ratios (e.g., above 50\%) prevent the model from learning meaningful representations~\cite{jiang2018mentornet}. 
Therefore, we select three representative noise settings of 10\%, 20\%, and 50\% for our experiments, 
while keeping the test set noise-free to ensure fair evaluation~\cite{li2024noisy}.  
\subsection{Baseline} 
We adopt a ResNet-based MinkowskiEngine~\cite{choy20194d} as the evaluation backbone, 
where the performance on unseen datasets reflects both segmentation quality and generalization ability. 
Building on this backbone, we are the first, to the best of our knowledge, to successfully overcome the challenges of transferring three representative noise-robust methods from the image domain to 3D LiDAR semantic segmentation and establish them as baselines:
\textbf{TCL}~\cite{bib:tcl}
applies Gaussian modeling and pseudo-labeling to enable contrastive learning. 
In the 3D point cloud domain, it suffers from two issues: 
(1) clustering-based clean sample selection is prohibitively expensive for large-scale point cloud data and significantly slows down inference; 
and (2) stochastic augmentations and varying point counts hinder stable prototype construction for pseudo-labels. 
We mitigate these by restricting contrastive learning to a small subset of aligned points and a limited number of mixed-view pairs.  \textbf{DISC}~\cite{bib:DISC}
dynamically selects clean samples while exploiting informative signals from noisy ones. 
However, it also faces challenges in large-scale point cloud segmentation: 
(1) the threshold-based comparison for distinguishing clean and hard labels is computationally expensive; 
and (2) points are not strictly aligned across batches, as newly appearing and disappearing points can drastically affect the statistics. 
To mitigate these issues, we only maintain statistics for points that reappear across scans, which stabilizes the estimation while reducing computational cost.  
\textbf{NPN}~\cite{bib:npn}
accumulates prediction statistics to construct partial and negative label sets, thereby mitigating the effect of overconfident noise. 
Its major challenge in 3D point cloud segmentation is that the varying number of points prevents the construction of consistent candidate label sets for individual samples, 
which makes it difficult to continuously update the candidates. 
To overcome this limitation, we instead take the maximum prediction within each batch as the candidate label, while still preserving complementary labels and the associated penalty function.    
\subsection{Dataset}
\textbf{SemanticKITTI~\cite{behley2019semantickitti}} is a large-scale LiDAR semantic segmentation benchmark with fine-grained point-wise annotations across diverse driving scenarios. 
It contains over 40k scans with approximately 100k points per frame, making it a comprehensive source domain for studying segmentation performance.  \textbf{NuScenes~\cite{caesar2020nuscenes}} is a multi-sensor autonomous driving dataset covering 1,000 driving scenes in diverse urban conditions with varying weather, traffic density, and viewpoints. 
Its LiDAR scans introduce significant domain shifts compared to the source and are used as an unseen target domain for evaluating cross-dataset generalization.  
\textbf{SemanticPOSS~\cite{pan2020poss}} is collected in Chinese urban environments and provides dense semantic annotations for street scenes under different geographic and environmental conditions. 
It serves as another target domain, complementing nuScenes by adding geographic and annotation diversity.  

This setup enables a joint study of in-domain segmentation on SemanticKITTI and cross-domain generalization on nuScenes and SemanticPOSS.  
\subsection{Evaluation Metric}
We follow the evaluation protocol of the Single Domain Generalization for LiDAR Semantic Segmentation (DGLSS)~\cite{kim2023single} baseline. Specifically, mean Intersection-over-Union (mIoU) is used to measure semantic segmentation performance on individual datasets, while the Arithmetic Mean (AM) and Harmonic Mean (HM) are employed to summarize results across multiple target domains. MIoU reflects segmentation accuracy, AM indicates the overall level of cross-domain generalization, and HM provides a stricter criterion by emphasizing performance balance across domains. Under this protocol, we first evaluate the DGLSS baseline at different noise ratios and observe that its mIoU drops significantly as noise increases, indicating degraded segmentation accuracy and generalization capability. We then compare representative noise-robust methods transferred from the image domain with our proposed dual-view framework - \textbf{DuNe} under the same setting. Higher mIoU corresponds to better segmentation performance, while larger AM and HM indicate stronger cross-domain generalization and robustness.

\section{Methodology}
\subsection{Problem Definition}

In this work, we study Domain Generalization for 3D LiDAR Semantic Segmentation under Noisy Labels, \ie, DGLSS-NL. Given a 3D point cloud $P = \{ p_i \in \mathbb{R}^3 \}_{i=1}^N$, the goal is to learn a segmentation model $f(\cdot)$
\begin{equation}
    y_i = f(p_i), \quad
    f:\mathbb{R}^3 \rightarrow \{1,...,C\}
\end{equation}
that assigns each point $p_i$ a true semantic label $y_i \in {1,\dots,C}$, where $C$ denotes the number of classes.

To enable DGLSS-NL, training is conducted on a noisy source domain $\mathcal{D}_S$ and evaluation is performed on an unseen target domain $\mathcal{D}_T$:
\begin{equation}
\mathcal{D}_S = \{(p_i, \tilde{y}_i)\}_{i=1}^{N_S},
\;
\mathcal{D}_T = \{p_j\}_{j=1}^{N_T},
\;
\mathcal{D}_S \cap \mathcal{D}_T = \emptyset,
\end{equation}

where $N_S, N_T$ are the number of points in the source and target domains, respectively and $\tilde{y}_i$ is defined as:
\begin{equation}
\tilde{y}_i = \begin{cases}
    y_i, & \text{with probability } (1-\eta) \\
    c \sim \mathcal{U}(\mathcal{C} \setminus \{y_i\}), & \text{with probability } \eta
\end{cases}
\end{equation}
where $\eta \in \{0.1, 0.2, 0.5\}$ is the symmetric noise ratio and $\mathcal{U}$ denotes uniform distribution over classes excluding $y_i$. The noise is synthetically injected to approximate real-world annotation errors (see Section~\ref{sec:nl}).

The training objective is to minimize the segmentation loss $\mathcal{L}_{\mathrm{seg}}$ on $\mathcal{D}_S$ under noisy supervision,
\begin{equation}
\min_{\theta}\; \mathbb{E}_{(p,\tilde{y}) \sim \mathcal{D}_S}\!\big[\mathcal{L}_{\mathrm{seg}}(f(p), \tilde{y})\big],
\end{equation}
while ensuring that $f(\cdot)$ is robust to label noise and can generalize effectively to the unseen domain $\mathcal{D}_T$.

\subsection{Dual-View Framework (DuNe)}

As illustrated in Fig.~\ref{fig:placeholder2}, each point cloud $P=\{p_i\}_{i=1}^N$ is first augmented into two complementary views using the PolarMix strategy~\cite{xiao2022polarmix}:
\begin{equation}
P^w = \{p_i^w\}_{i=1}^N, \qquad
P^s = \{p_i^s\}_{i=1}^{N'},
\end{equation}
where $P^w$ (weak view) preserves the structural fidelity of the original scan with the same cardinality $N$, and $P^s$ (strong view) may have $N'\geq N$ points due to additional rotated-pasted instances introduced by PolarMix.
In practice, PolarMix applies a scene-level swapping $Sw(\cdot)$ and an instance-level rotate-paste $Rp(\cdot)$ operation, yielding
\begin{equation}
P^s = Sw(P, P') \oplus Rp(P, P'),
\end{equation}
where $P'$ denotes another scan, and $\oplus$ indicates concatenation.

Subsequently, both views are further processed by the baseline DGLSS module~\cite{kim2023single} to enhance sparsity. Specifically, the 3D point cloud is projected into a range view, and a randomly selected row is removed to simulate beam-missing artifacts:
\begin{equation}
\mathcal{A}(P) = \text{RowDrop}(\text{RangeProj}(P)),
\end{equation}
which reduces the number of points to $\tilde{N}<N$ (or $\tilde{N}<N'$ for strong views).
This yields four derived views:
\begin{align}
P^{ss} = \mathcal{A}(P^s) \in \mathbb{R}^{\tilde{N}' \times 3};
\quad P^{sa} = P^s \in \mathbb{R}^{N' \times 3}; \\
P^{ws} = \mathcal{A}(P^w) \in \mathbb{R}^{\tilde{N} \times 3}; 
\quad P^{wa} = P^w \in \mathbb{R}^{N \times 3}.
\end{align}

Each view is voxelized and encoded with a ResNet-based sparse convolutional network implemented via MinkowskiEngine~\cite{choy20194d}:
\begin{equation}
F^v = \phi(P^v), \qquad v \in \{ss, sa, ws, wa\},
\end{equation}
where $\phi(\cdot)$ denotes the sparse 3D encoder that maps voxelized point clouds into high-dimensional feature tensors. 

A lightweight decoder with transposed sparse convolutions is then applied to upsample the encoded features back to the original voxel resolution for point-wise prediction. In the strong-view branch, the decoded features are concatenated and directly classified as
\begin{equation}
\hat{y}^s = \psi\big(\mathrm{Concat}(F^{ss}, F^{sa})\big),
\end{equation}
while in the weak-view branch, the concatenated features are decoded and split back into two parts:
\begin{equation}
(\hat{y}^{ws}, \hat{y}^{wa}) = \psi\big(\mathrm{Concat}(F^{ws}, F^{wa})\big),
\end{equation}
where $\psi(\cdot)$ denotes the task-specific decoder and classifier. During inference, only the strong branch is utilised, while the weak branch and consistency losses are disabled for computational efficiency.

\subsection{Loss Function}
To improve robustness against sparsity variation, semantic ambiguity, and noisy supervision, we adopt two complementary objectives: DGLSS module and NPN module. 

\textbf{DGLSS Loss}: Following the DGLSS framework~\cite{kim2023single}, we introduce two consistency terms in addition to the standard cross-entropy (CE) loss.
First, the Sparsity-Invariant Feature Consistency (SIFC) encourages feature alignment across scans with different sparsity levels:
\begin{equation}
\mathcal{L}_{\text{SIFC}} = \frac{1}{N} \sum_{i=1}^N \| F^{ws}_i - F^{wa}_i \|_1 .
\end{equation}

Second, the Semantic Correlation Consistency (SCC) enforces stable inter-class relationships across domains. For each scan, decoded features are passed through a metric learner to obtain class-wise prototypes $Z_i \in \mathbb{R}^{C \times d}$, where each row is the average embedding of a class. The SCC loss aligns pairwise prototype correlations across scans:
\begin{equation}
\mathcal{L}_{\text{SCC}} = \frac{1}{L} \sum_{i} \sum_{j \neq i} 
\big( Z_i^\top Z_i - Z_j^\top Z_j \big),
\end{equation}

Finally, besides these two consistency terms, we also adopt a weighted CE loss $\mathcal{L}_{\text{sem}}$ to handle class imbalance. The overall DGLSS objective is
\begin{equation}
\mathcal{L}_{\text{DGLSS}} = \mathcal{L}_{\text{sem}} + \alpha \mathcal{L}_{\text{SIFC}} + \beta \mathcal{L}_{\text{SCC}} ,
\end{equation}
where $\alpha$ and $\beta$ balance the contributions.

\textbf{NPN Loss}: To further handle noisy labels, we adopt NPN~\cite{bib:npn}. Given the predicted label $\hat{y}^s$ from the strong branch, we decompose the label space into a candidate label set $\hat{Y}$ (including $\hat{y}^{ws}$, $\hat{y}^{wa}$ and $\hat{y}^s$) and its complementary label set (all remaining classes).
The Partial Label Learning (PLL) term encourages the model to predict one of the candidate labels, while the Negative Learning (NL) term explicitly penalizes complementary labels:
\begin{equation}
L_{\text{NL}} = - \frac{1}{N} \sum_{i=1}^N \sum_{c \neq \hat{y}_i} \log \big( 1 - p_\theta(\hat{y}=c \mid x_i) \big), \; \hat{y}\in \hat{Y}.
\end{equation}
The final NPN objective integrates PLL, NL, and a confidence regularization penalty:
\begin{equation}
\mathcal{L}_{\text{NPN}} = \mu \, L_{\text{NL}} + \nu \, L_{\text{CE}} + L_{\text{pen}},
\end{equation}
where $\mu$ and $\nu$ are balancing coefficients.

\textbf{Overall Objective}: Finally, we integrate the DGLSS and NPN losses within the dual-view framework.  
In addition to the consistency and semantic correlation regularizations, 
we introduce a dual-view feature consistency loss $\mathcal{L}_{\text{FC}}$ to encourage similarity between the strong and weak view representations. The overall training objective is defined as
\begin{equation}
\mathcal{L}_{\text{total}} = \mathcal{L}_{\text{DGLSS}} + \mathcal{L}_{\text{NPN}} + \lambda \, \mathcal{L}_{\text{FC}},
\end{equation}
where $\lambda$ balances the dual-view feature consistency term.  
This unified design jointly enforces feature-level consistency, semantic correlation, and noise-robust supervision, 
enabling the network to learn reliable representations from noisy labels while maintaining strong generalization across domains.

\section{Experiments}

\subsection{Experimental Setup}
All models are implemented in PyTorch with sparse convolution operations from MinkowskiEngine, and trained on NVIDIA A100 GPUs with 40GB memory.

\noindent\textbf{Backbone.} We adopt the DGLSS~\cite{kim2023single} model as the backbone baseline and evaluate its performance under four supervision settings: 
(i) clean labels (noise-free baseline), 
(ii) noisy labels without correction, 
(iii) three noise-robust strategies migrated from the image domain (TCL~\cite{bib:tcl}, DISC~\cite{bib:DISC}, and NPN~\cite{bib:npn}), 
and (iv) our proposed dual-view learning framework - \textbf{DuNe}, which incorporates bottleneck consistency and adopts partial/negative label supervision.  
For in-domain evaluation, models are trained and tested on the SemanticKITTI~\cite{behley2019semantickitti} dataset (K $\rightarrow$ K). 
For cross-domain evaluation, models are trained on SemanticKITTI and directly tested on nuScenes~\cite{caesar2020nuscenes} (K $\rightarrow$ N) and SemanticPOSS~\cite{pan2020poss} (K $\rightarrow$ P) without target-domain fine-tuning.

\noindent\textbf{Hyperparameters.} 
Stochastic gradient descent (SGD) with momentum 0.9 and weight decay $1e^{-4}$ is used as the optimizer.
The initial learning rate is set to 0.01 and decayed by a cosine annealing schedule.  
Unless otherwise specified, the loss balancing weights are set to $\mu = 1.0$ for the consistency loss, $\nu = 1.0$ for the negative cross-entropy loss, and $\lambda = 2.0$ for the prototype contrastive loss.
\begin{figure*}[!htbp]
    \centering
    \includegraphics[width=0.8\textwidth]{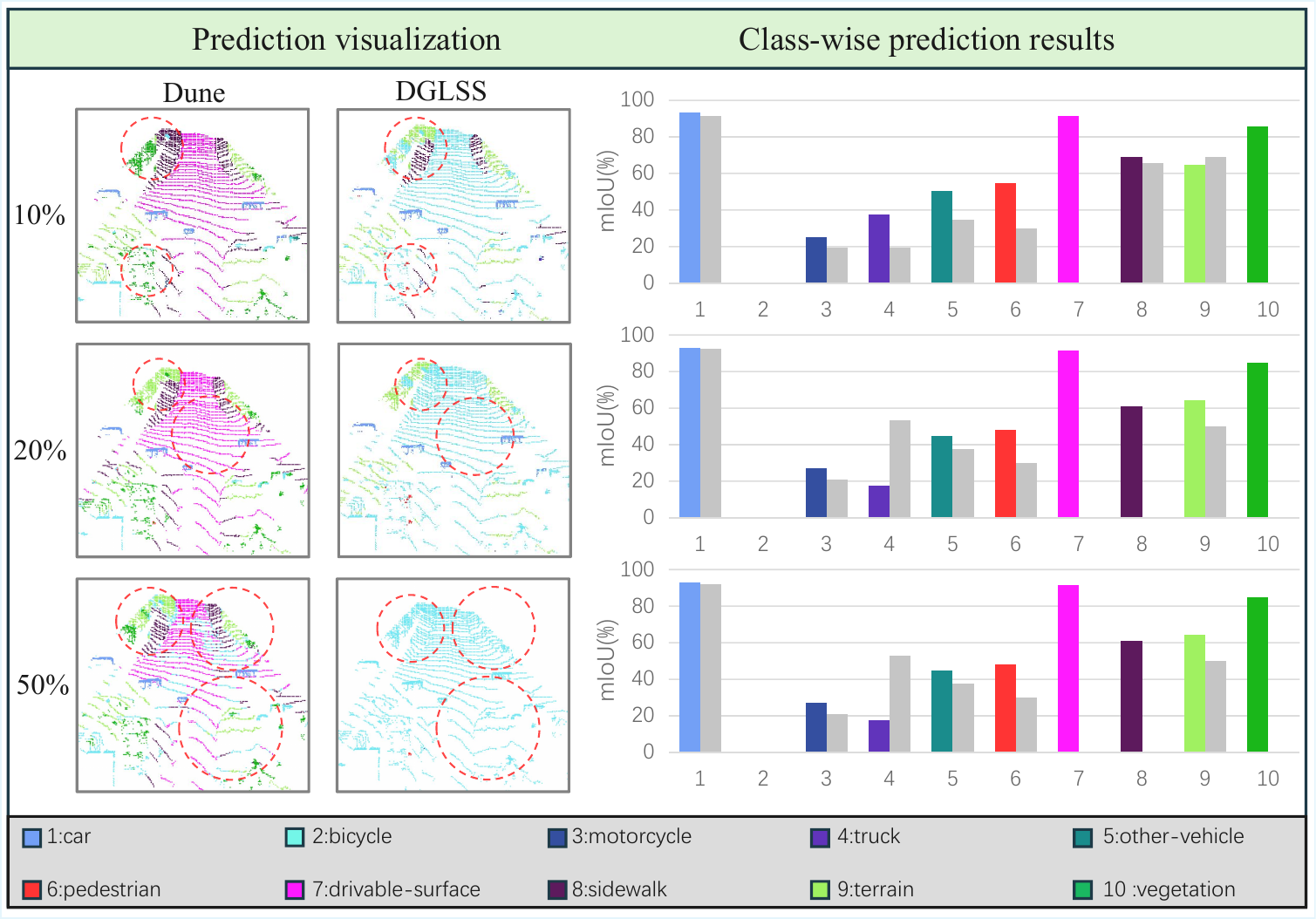}
    \caption{
    Qualitative results under 10\%, 20\%, and 50\% label noise. 
    For each noise ratio, we compare segmentation predictions of our framework DuNe with the baseline DGLSS. 
    Corresponding class-wise prediction histograms (10 categories) are also shown. 
    The left column visualizes predictions, where circles highlight differences between methods.
    The right column presents class-wise prediction distributions:  colored bars indicate results from our method, while gray bars correspond to predictions from the DGLSS baseline. 
    Our method produces more coherent segmentations and maintains balanced class distributions, even under severe noise.
    }
    \label{fig:qualitative}
\end{figure*}
\begin{table}[t]
\centering
\caption{Comparison of 3D semantic segmentation results trained on KITTI under different noise ratios. All values are mIoU(\%)($\uparrow$). 
K, N, and P denote SemanticKITTI, nuScenes, and SemanticPOSS.  K$\rightarrow$\textcolor{blue}{N} indicate models trained on K and evaluated on \textcolor{blue}{N}.}
\begin{tabular}{c|cccccc}
\toprule
\multicolumn{1}{c|}{\multirow{2.5}{*}{\textbf{Dataset}}} 
 & \multicolumn{6}{c}{\textbf{Symmetric Noise}} \\
\cline{2-7} 
& \rule{0pt}{3ex}\textbf{0\%} & \textbf{2\%} & \textbf{5\%} & \textbf{10\%} & \textbf{20\%} & \textbf{50\%} \\
\midrule
K $\rightarrow$ \textcolor{blue}{K}   & 58.06 & 55.70 & 54.75 & 32.99 & 28.35 & 10.86 \\
K $\rightarrow$ \textcolor{blue}{N}   & 42.28 & 34.96 & 36.42 & 21.76 & 16.48 & 7.53  \\
K $\rightarrow$ \textcolor{blue}{P}   & 49.09 & 41.81 & 44.28 & 22.29 & 18.86 & 9.38  \\
\midrule
AM & 49.81 & 45.33 & 44.28 & 27.38 & 22.42 & 9.19 \\
HM & 48.99 & 42.96 & 43.74 & 26.23 & 20.85 & 9.25 \\
\bottomrule
\end{tabular}
\label{tab:noise_results}
\vskip-3ex
\end{table}

\subsection{Results}

\noindent\textbf{Effect of Label Noise on the Baseline.}  
Table~\ref{tab:noise_results} shows that the baseline DGLSS is highly vulnerable to noisy supervision. 
With clean labels, it achieves 58.06\% mIoU on SemanticKITTI, 42.28\% on nuScenes, and 49.09\% on SemanticPOSS. 
However, performance deteriorates rapidly with increasing noise: on SemanticKITTI, mIoU drops to 32.99\% at 10\% noise and to only 10.86\% at 50\%. 
Similar trends appear in cross-dataset evaluations, where nuScenes decreases from 42.28\% to 7.53\% and SemanticPOSS from 49.09\% to 9.38\%.  
While low noise levels (2\%–5\%) have only a marginal effect due to the inherent stability of the backbone, moderate to high noise ratios (20\%–50\%) substantially impair both in-domain accuracy and cross-domain robustness, confirming the necessity of noise-robust strategies in LiDAR semantic segmentation.  
\begin{table}[t!]
\centering
\caption{Comparison of methods under different noise ratios. All values are mIoU (\%)($\uparrow$). 
K, N, and P denote SemanticKITTI, nuScenes, and SemanticPOSS, respectively.
All results are obtained by training on SemanticKITTI and evaluating on the datasets indicated below.}
\begin{tabular}{c|c|ccccc}
\toprule
\textbf{Noise Ratio} & \textbf{Method} & \textbf{K} & \textbf{N} & \textbf{P} & \textbf{AM} & \textbf{HM} \\
\midrule
\multirow{4}{*}{10\%} 
 & TCL~\cite{bib:tcl}  & 19.59 & 18.98 & 22.11 & 19.28 & 19.28 \\
 & DISC~\cite{bib:DISC} & 41.42 & 36.79 & 31.03 & 36.23 & 35.48 \\
 & NPN~\cite{bib:npn}  & 52.05 & 38.00 & 42.88 & 45.02 & 43.93 \\
 & DuNe & \textbf{56.86} & \textbf{42.28} & \textbf{52.58} & \textbf{49.57} & \textbf{48.50} \\
\midrule
\multirow{4}{*}{20\%} 
 & TCL  & 14.14 & 10.33 &  8.86 & 10.82 & 10.80 \\
 & DISC & 36.68 & 34.44 & 25.85 & 31.26 & 30.33 \\
 & NPN  & 48.72 & 31.84 & 36.63 & 40.28 & 38.53 \\
 & DuNe & \textbf{53.20} & \textbf{39.69} & \textbf{48.05} & \textbf{46.44} & \textbf{45.46} \\
\midrule
\multirow{4}{*}{50\%} 
 & TCL  & 10.37 & 11.28 & 10.19 & 10.82 & 10.82 \\
 & DISC & 33.64 & 28.22 & 24.98 & 29.31 & 28.67 \\
 & NPN  & 41.78 & 30.59 & 31.07 & 36.19 & 35.32 \\
 & DuNe & \textbf{52.37} & \textbf{37.18} & \textbf{43.07} & \textbf{44.78} & \textbf{43.49} \\
\bottomrule
\end{tabular}
\label{tab:noise_comparison}
\vskip-3ex
\end{table}

\noindent\textbf{Comparison with Transferred Noise-Robust Methods.}  Table~\ref{tab:noise_comparison} compares \textbf{DuNe} with three representative noise-robust strategies adapted from the image domain: TCL~\cite{bib:tcl}, DISC~\cite{bib:DISC}, and NPN~\cite{bib:npn}.  
TCL consistently fails to improve robustness and in many cases even underperforms the noisy baseline, showing that this method does not transfer effectively to 3D LiDAR segmentation.  
DISC achieves moderate gains by dynamically selecting cleaner samples, e.g., on SemanticKITTI it improves performance from 32.99\% to 41.42\% at 10\% noise, with similar trends on nuScenes and SemanticPOSS.  
NPN further enhances robustness by leveraging partial and negative labels, reaching 41.78\% on SemanticKITTI, 30.59\% on nuScenes, and 31.07\% on SemanticPOSS at 50\% noise.  

\begin{table*}[t!]
\centering
\caption{Per-class IoU(\%)($\uparrow$) under different noise ratios. 
Here, \textbf{K}, \textbf{N}, and \textbf{P} denote SemanticKITTI, nuScenes, and SemanticPOSS, respectively. 
K$\rightarrow$\textcolor{blue}{N} indicate models trained on K and evaluated on \textcolor{blue}{N}. We omit the non-existent common classes in SemanticPOSS and mark them with ‘-’.}
\resizebox{\textwidth}{!}{
\begin{tabular}{c|l|l|c|
c|c|c|c|c|c|c|c|c|c|c|c}
\toprule
Noise Ratio & Method & Dataset & mIoU &
\rotatebox{90}{\sqrgb{0.39}{0.59}{1.00} car} &
\rotatebox{90}{\sqrgb{0.59}{1.00}{1.00} bicycle} &
\rotatebox{90}{\sqrgb{0.24}{0.35}{0.71} motorcycle} &
\rotatebox{90}{\sqrgb{0.47}{0.24}{0.71} truck} &
\rotatebox{90}{\sqrgb{0.00}{0.59}{0.59} other-vehicle} &
\rotatebox{90}{\sqrgb{1.00}{0.24}{0.24} pedestrian} &
\rotatebox{90}{\sqrgb{1.00}{0.00}{1.00} drivable-surface} &
\rotatebox{90}{\sqrgb{0.47}{0.00}{0.47} sidewalk} &
\rotatebox{90}{\sqrgb{0.71}{1.00}{0.47} terrain} &
\rotatebox{90}{\sqrgb{0.00}{0.71}{0.00} vegetation} &
AM & HM \\
\midrule
10\% & DGLSS& \textbf{K $\rightarrow$ \textcolor{blue}{K}} & 32.99 & 91.44 & 0.08 & 19.38 & 19.63 & 34.53 & 30.13 & 0.01  & \textbf{65.72} & \textbf{69.00} & 0.00  & 27.38 & 26.23 \\
     &      & \textbf{K $\rightarrow$ \textcolor{blue}{N}} & 21.76 & 76.06 & 0.02 & 17.51 & 11.94 & 17.65 & 24.46 & 0.01  & 35.48 & 34.50 & 0.00  &       &       \\
     &      & \textbf{K $\rightarrow$ \textcolor{blue}{P}} & 22.29 & 59.66 & 4.60 & -  & -  & -  & 45.68 & 0.01  & 0.00  & 1.49  & -  &       &       \\ 
    \cline{3-16}
& \rule{0pt}{3ex}\hspace{0.1em}DuNe & \textbf{K $\rightarrow$ \textcolor{blue}{K}}    
& \textbf{56.86} & \textbf{92.92} & 0.58 & 13.24 & \textbf{54.59} & \textbf{51.14} & \textbf{50.78} & \textbf{91.70} & 62.19 & 68.25 & \textbf{83.19} & \textbf{49.57} &  \textbf{48.50}   \\

&      & \textbf{K $\rightarrow$ \textcolor{blue}{N}} 
& 42.28 & 78.06 & 0.11 & \textbf{33.15} & 26.76 & 12.33 & 44.34 & 84.98 & 34.48 & 41.56 & 67.01 &       &       \\

&      & \textbf{K $\rightarrow$ \textcolor{blue}{P}} 
& 52.58 & 64.53 & \textbf{7.52} & -     & -     & -     & 50.61 & 66.21 & -     & 74.03 & -     &       &       \\
\midrule
20\% &DGLSS& \textbf{K $\rightarrow$ \textcolor{blue}{K}} & 28.35 & 92.24 & 0.06 & 20.82 & \textbf{53.12} & 37.57 & 29.85 & 0.00  & 0.00  & 49.84 & 0.00  & 22.42 & 20.85 \\ 
     &      & \textbf{K $\rightarrow$ \textcolor{blue}{N}} & 16.48 & 73.26 & 0.02 & 15.11 & 18.07 & 7.79  & 29.91 & 0.00  & 0.00  & 20.67 & 0.00  &       &       \\
     &      & \textbf{K $\rightarrow$ \textcolor{blue}{P}} & 18.66 & 51.84 & 4.58 & -  & -  & -  & 35.38 & 0.00  & 0.00  & 1.52  & -  &       &       \\ 
     \cline{3-16}
     & \rule{0pt}{3ex}\hspace{0.1em}DuNe & \textbf{K $\rightarrow$ \textcolor{blue}{K}} & \textbf{53.20} & \textbf{92.76} & 0.28 & 27.19 & 17.46 & \textbf{44.79} & \textbf{47.92} & \textbf{91.32} & \textbf{60.83} & 64.45 & \textbf{84.99} & 46.44 & 45.46 \\
     &      & \textbf{K $\rightarrow$ \textcolor{blue}{N}} & 39.69 & 79.61 & 0.08 & \textbf{36.30} & 15.20 & 15.83 & 38.08 & 83.70 & 36.58 & 35.48 & 56.05 &       &       \\
     &      & \textbf{K $\rightarrow$ \textcolor{blue}{P}} & 48.05 & 61.86 & \textbf{6.64} & -  & -  & -  & 43.25 & 60.43 & -  & \textbf{68.05} & -  &       &       \\ 
\midrule
50\% & DGLSS& \textbf{K $\rightarrow$ \textcolor{blue}{K}} & 10.86 & 0.38  & 0.06 & 8.24  & 21.47 & \textbf{45.74} & 32.68 & 0.00  & 0.00  & 0.00  & 0.00  & 9.19  & 8.89  \\
     &      & \textbf{K $\rightarrow$ \textcolor{blue}{N}} & 7.53  & 0.13  & 0.02 & 13.82 & 23.84 & 8.02  & 29.44 & 0.00  & 0.00  & 0.00  & 0.00  &       &       \\
     &      & \textbf{K $\rightarrow$ \textcolor{blue}{P}} & 9.38  & 0.05  & 4.38 & -  & -  & -  & 42.47 & 0.00  & 0.00  & 0.00  & -  &       &       \\
     \cline{3-16}
     & \rule{0pt}{3ex}\hspace{0.1em}DuNe & \textbf{K $\rightarrow$ \textcolor{blue}{K}} & \textbf{52.37} & \textbf{90.05} & 0.22 & 19.33 & \textbf{41.72} & 36.48 & \textbf{42.78} & \textbf{90.84} & \textbf{61.24} & 59.62 & \textbf{81.46} & 44.78 & 43.49 \\
     &      & \textbf{K $\rightarrow$ \textcolor{blue}{N}} & 37.18 & 72.60 & 0.06 & \textbf{31.09} & 17.45 & 12.24 & 35.07 & 78.33 & 38.55 & 25.22 & 61.21 &       &       \\
     &      & \textbf{K $\rightarrow$ \textcolor{blue}{P}} & 43.07 & 54.58 & \textbf{6.65} & -  & -  & -  & 40.46 & 40.17 & -  & \textbf{73.48} & -  &       &       \\
\bottomrule
\end{tabular}}
\label{tab:results_per_class}
\end{table*}

\begin{table}
\centering
\setlength{\tabcolsep}{1.5pt}
\caption{Ablation study on the contributions of NPN and PolarMix under the DGLSS setting. 
All values are reported as mIoU(\%)($\uparrow$). 
Here, \textbf{K}, \textbf{N}, and \textbf{P} denote SemanticKITTI, nuScenes, and SemanticPOSS, respectively. 
\textbf{K$\rightarrow$\textcolor{blue}{N}} indicate models trained on SemanticKITTI and evaluated on nuScenes.}
{\begin{tabular}{l|c|c|c|c|c}
\toprule
\textbf{Method} & \textbf{K $\rightarrow$ \textcolor{blue}{K}} & \textbf{K $\rightarrow$ \textcolor{blue}{N}} & \textbf{K $\rightarrow$ \textcolor{blue}{P}}  & \textbf{AM}  & \textbf{HM} \\
\midrule                         
10\% DGLSS & 32.99 & 21.76 & 22.29 & 27.38 & 26.23 \\ 
10\% DGLSS + PolarMix       & 44.12 & 33.48 & 43.94 & 38.80 & 38.07 \\
10\% DGLSS + NPN    & 52.05 & 38.00 & 42.88 & 45.02 & 43.93 \\
10\% DGLSS + NPN + PolarMix    & 55.75 & 35.92 & 38.67 & 45.83 & 43.69 \\
\midrule 
50\% Noise strong branch & 30.71  & 19.84  & 24.90  & 25.27  & 24.11 \\
20\% Noise weak branch   & 51.03  & 33.74  & 42.86  & 42.38  & 40.62 \\
10\% Noise weak branch   & 54.13  & 36.96  & 48.53  & 45.55  & 43.93 \\
10\% DuNe                & \textbf{56.86} & \textbf{42.28} & \textbf{52.58} & \textbf{49.57} & \textbf{48.50} \\                        
\bottomrule
\end{tabular}}
\label{tab:ablation}
\vskip-3ex
\end{table}

\noindent\textbf{Performance of the Proposed Framework.}  Our method \textbf{DuNe} consistently achieves the best results across all datasets and noise levels according to Table~\ref{tab:results_per_class}.  
It maintains 52.37\% on SemanticKITTI, 37.18\% on nuScenes, and 43.07\% on SemanticPOSS under 50\% noise, substantially outperforming TCL and DISC and clearly surpassing NPN.  
These results demonstrate that while generic 2D noise-robust methods provide partial benefits, principled designs tailored to 3D LiDAR are essential for mitigating the impact of noisy supervision.
More importantly, our approach enhances generalization to unseen domains. 
Under 10\% symmetric noise, when trained on SemanticKITTI, our method achieves 56.86\% mIoU on SemanticKITTI, 42.28\% mIoU on nuScenes and 52.58\% on SemanticPOSS, resulting in AM and HM of 49.57\% and 48.50\%, respectively. These results consistently surpass the baseline and approach the performance of DGLSS trained with clean labels, demonstrating strong robustness against both label noise and domain shift. Overall, the proposed dual-view learning framework not only restores segmentation accuracy under noisy supervision but also substantially improves cross-domain generalization, establishing a solid benchmark for noise-robust LiDAR semantic segmentation.
\subsection{Ablation Study}
Table~\ref{tab:ablation} reports the ablation results on the contributions of NPN and PolarMix under the DGLSS setting. 
Several important observations can be made. 

\noindent\textbf{Effect of PolarMix.}  
Compared to the plain DGLSS baseline (32.99\%, 21.76\%, and 22.29\% on SemanticKITTI, nuScenes, and SemanticPOSS), adding PolarMix under 10\% label noise already yields clear gains, reaching 44.12\%, 33.48\%, and 43.94\%.  
This indicates that geometric mixing can effectively enhance data diversity and support cross-domain transfer, although the in-domain improvement remains moderate.  

\noindent\textbf{Effect of NPN.}  
Applying NPN instead of PolarMix results in more substantial robustness improvements, with performance rising to 52.05\%, 38.00\%, and 42.88\% on the three benchmarks.  
This demonstrates that partial-label and negative learning significantly mitigate the negative effect of corrupted annotations and stabilize the optimization.  

\noindent\textbf{Combination of NPN and PolarMix.}  
When both are applied simultaneously, the model achieves 55.75\%, 35.92\%, and 38.67\%.  
Although the SemanticPOSS result is slightly lower than NPN-only, the SemanticKITTI accuracy surpasses both single variants.  
These results suggest that noise modeling and geometric augmentation are complementary, each contributing to robustness and generalization under noisy supervision.  

\noindent\textbf{Effect of dual-branch consistency.}  
Although combining NPN and PolarMix improves performance, the overall generalization remains limited, with AM and HM reaching only 45.83\% and 43.69\%.
To further bridge the semantic gap between different representations, we introduce a consistency loss on the bottleneck features of the dual-branch design.
Our experiments show that under 10\% and 20\% noise, using the strong achieves noticeably performance of generalization with higher AM and HM. This suggests that the denser augmented point clouds provide richer cues and promote the learning of transferable knowledge.
However, under 50\% noise,  the dense but corrupted strong set amplifies the adverse effects of label noise, causing performance to collapse. Therefore, we adopt a selective strategy: using the strong set for 10\% and 20\% noise, and the weak set for 50\% noise, enabling more robust performance across different noise levels.

\noindent\textbf{Full framework.} 
Our method \textbf{DuNe} achieves the best results across all datasets, with 56.86\%, 42.28\%, and 52.58\% on SemanticKITTI, nuScenes, and SemanticPOSS under 10\% noise, respectively. 
These improvements confirm that integrating NPN and PolarMix within our dual-view training strategy is crucial for achieving robustness and generalization under noisy supervision.

\section{Conclusion}

We studied robust domain-generalized LiDAR semantic segmentation under noisy supervision. To establish a benchmark, we introduced symmetric label noise into SemanticKITTI, nuScenes, and SemanticPOSS, and adapted three representative noise-robust strategies from the image domain. This revealed both the sensitivity of standard training to corrupted labels and the limited transferability of 2D methods to 3D point clouds.
To address this, we proposed \textbf{DuNe}, a dual-view learning framework combining PolarMix augmentation, bottleneck consistency, and partial/negative label supervision. Experiments showed that \textbf{DuNe} not only recovers segmentation accuracy under noise but also improves cross-dataset generalization. We hope this benchmark and framework foster future research on noise-robust LiDAR perception for autonomous driving.

\bibliography{my}

@STRING{IEEE_J_ITS        = "{IEEE} Trans. Intell. Transp. Syst."}

@String{CVPR = "Proc. IEEE Conf. Comput. Vis. Pattern Recognit."}

@String{ICCV = "Proc. IEEE Int. Conf. Comput. Vis."}

@String{ECCV = "Proc. Eur. Conf. Comput. Vis."}

@String{ICRA = "Proc. IEEE Int. Conf. Robot. Autom."}

@String{IROS = "Proc. IEEE/RSJ Int. Conf. Intell. Robots Syst."}

@String{NeurIPS = "Adv. Neural Inf. Process. Syst."}

@String{ICML = "Int. Conf. Mach. Learn."}

@String{AAAI = "Proc. AAAI Conf. Artif. Intell."}

@String{ICPR = "Proc. Int. Conf. Pattern Recognit."}

@article{bib:lidar-survey,
  title={A Review of Point Cloud Registration Algorithms for Mobile Robotics},
  author={Pomerleau, Fran{\c{c}}ois and Colas, Francis and Siegwart, Roland},
  journal={Foundations and Trends{\textregistered} in Robotics},
  volume={4},
  number={1},
  pages={1--104},
  year={2015},
  publisher={Now Publishers Boston—Delft}
}

@inproceedings{kim2023single,
  title={Single Domain Generalization for LiDAR Semantic Segmentation},
  author={Kim, Hyeonseong and Kang, Yoonsu and Oh, Changgyoon and Yoon, Kuk-Jin},
  booktitle=CVPR,
  pages={17587--17598},
  year={2023}
}

@article{peng2022mass,
  title={MASS: Multi-attentional semantic segmentation of LiDAR data for dense top-view understanding},
  author={Peng, Kunyu and Fei, Juncong and Yang, Kailun and Roitberg, Alina and Zhang, Jiaming and Bieder, Frank and Heidenreich, Philipp and Stiller, Christoph and Stiefelhagen, Rainer},
  journal=IEEE_J_ITS,
  volume={23},
  number={9},
  pages={15824--15840},
  year={2022},
  publisher={IEEE}
}

@inproceedings{bib:waymo,
  author    = {Sun, Pei and Kretzschmar, Henrik and Dotiwalla, Xerxes and Chouard, Aurelien and others},
  title     = {Scalability in Perception for Autonomous Driving: Waymo Open Dataset},
  booktitle = CVPR,
  pages     = {2443--2451},
  year      = {2020}
}

@article{bib:label-noise-survey,
  author    = {Frenay, Benoit and Verleysen, Michel},
  title     = {Classification in the Presence of Label Noise: A Survey},
  journal   = {IEEE Transactions on Neural Networks and Learning Systems},
  volume    = {25},
  number    = {5},
  pages     = {845--869},
  year      = {2014}
}

@inproceedings{bib:tcl,
  title={Twin Contrastive Learning with Noisy Labels},
  author={Huang, Zhizhong and Zhang, Junping and Shan, Hongming},
  booktitle=CVPR,
  year={2023}
}

@inproceedings{bib:disc,
    author    = {Li, Yifan and Han, Hu and Shan, Shiguang and Chen, Xilin},
    title     = {DISC: Learning From Noisy Labels via Dynamic Instance-Specific Selection and Correction},
    booktitle = CVPR,
    month     = {June},
    year      = {2023},
    pages     = {24070-24079}
}

@inproceedings{bib:npn,
  author    = {M. Sheng and Z. Sun and Z. Cai and T. Chen and Y. Zhou and Y. Yao},
  title     = {Adaptive Integration of Partial Label Learning and Negative Learning for Enhanced Noisy Label Learning},
  booktitle = AAAI,
  volume    = {38},
  number    = {5},
  pages     = {4820--4828},
  month     = {Mar.},
  year      = {2024}
}

@inproceedings{bib:PointNet,
  author    = {Qi, Charles R and Su, Hao and Mo, Kaichun and Guibas, Leonidas J},
  title     = {PointNet: Deep Learning on Point Sets for 3D Classification and Segmentation},
  booktitle = CVPR,
  pages     = {652--660},
  year      = {2017}
}

@inproceedings{fei2021pillarsegnet,
  title={Pillarsegnet: Pillar-based semantic grid map estimation using sparse lidar data},
  author={Fei, Juncong and Peng, Kunyu and Heidenreich, Philipp and Bieder, Frank and Stiller, Christoph},
  booktitle={2021 IEEE intelligent vehicles symposium (IV)},
  pages={838--844},
  year={2021},
  organization={IEEE}
}

@inproceedings{wu2018squeezeseg,
  title={SqueezeSeg: Convolutional neural nets with recurrent CRF for real-time road-object segmentation from 3D LiDAR point cloud},
  author={Wu, Bichen and Wan, Alvin and Yue, Xiangyu and Keutzer, Kurt},
  booktitle=ICRA,
  year={2018}
}

@inproceedings{qi2017pointnetplusplus,
  title={PointNet++: Deep hierarchical feature learning on point sets in a metric space},
  author={Qi, Charles R and Yi, Li and Su, Hao and Guibas, Leonidas J},
  booktitle={NeurIPS},
  year={2017}
}

@inproceedings{milioto2019rangenet++,
  title={RangeNet++: Fast and accurate LiDAR semantic segmentation},
  author={Milioto, Andres and Vizzo, Ignacio and Behley, Jens and Stachniss, Cyrill},
  booktitle=IROS,
  year={2019}
}

@article{June3DSeg,
  title={Zero-Shot Detection Of Buildings In Mobile Lidar Using Language Vision Model},
  author={Moh Goo, June and Zeng, Zichao and Boehm, Jan},
  journal={ISPRS-International Archives of the Photogrammetry, Remote Sensing and Spatial Information Sciences},
  volume={48},
  pages={107--113},
  year={2024}
}

@inproceedings{zhou2018voxelnet,
  title={VoxelNet: End-to-end learning for point cloud based 3D object detection},
  author={Zhou, Yin and Tuzel, Oncel},
  booktitle=CVPR,
  year={2018}
}

@inproceedings{graham20183d,
  author    = {Benjamin Graham and Martin Engelcke and Laurens van~der Maaten},
  title     = {3D Semantic Segmentation with Submanifold Sparse Convolutional Networks},
  booktitle = CVPR,
  year      = {2018}
}

@inproceedings{hu2020randla,
  author    = {Qingyong Hu and Bo Yang and Linhai Xie and Stefan Rosa and Yulan Guo and Zhihua Wang and Niki Trigoni and Andrew Markham},
  title     = {RandLA-Net: Efficient Semantic Segmentation of Large-Scale Point Clouds},
  booktitle = CVPR,
  year      = {2020}
}

@inproceedings{caesar2020nuscenes,
  title={nuScenes: A multimodal dataset for autonomous driving},
  author={Holger Caesar and Varun Bankiti and Alex H. Lang and Sourabh Vora and Venice Erin Liong and Qiang Xu and Anush Krishnan and Yu Pan and Giancarlo Baldan and Oscar Beijbom},
  booktitle=CVPR,
  year={2020}
}

@inproceedings{huang2021metasets,
  title={Metasets: Meta-learning on point sets for generalizable representations},
  author={Huang, Chao and Cao, Zhangjie and Wang, Yunbo and Wang, Jianmin and Long, Mingsheng},
  booktitle={Proceedings of the IEEE/CVF Conference on Computer Vision and Pattern Recognition},
  pages={8863--8872},
  year={2021}
}

@inproceedings{huang2022manifold,
  title={Manifold adversarial learning for cross-domain 3d shape representation},
  author={Huang, Hao and Chen, Cheng and Fang, Yi},
  booktitle=ECCV,
  pages={272--289},
  year={2022},
  organization={Springer}
}

@inproceedings{patrini2017forward,
author = {Patrini, Giorgio and Rozza, Alessandro and Krishna Menon, Aditya and Nock, Richard and Qu, Lizhen},
title = {Making Deep Neural Networks Robust to Label Noise: A Loss Correction Approach},
booktitle = CVPR,
year = {2017}
}

@inproceedings{jiang2018mentornet,
  title={{MentorNet:} {Learning} Data-Driven Curriculum for Very Deep Neural Networks on Corrupted Labels},
  author={Jiang, Lu and Zhou, Zhengyuan and Leung, Thomas and Li, Li-Jia and Fei-Fei, Li},
  booktitle=ICML,
  pages={2304--2313},
  year={2018}
}

@inproceedings{bib:sample-selection,
  author    = {Han, Bo and Yao, Quanming and Yu, Xingrui and Niu, Gang and Xu, Miao and Hu, Weihua and Tsang, Ivor and Sugiyama, Masashi},
  title     = {Co-teaching: Robust Training of Deep Neural Networks with Extremely Noisy Labels},
  booktitle = NeurIPS,
  pages     = {8527--8537},
  year      = {2018}
}

@inproceedings{ghosh2017robust,
  title={Robust loss functions under label noise for deep neural networks},
  author={Ghosh, Aritra and Kumar, Himanshu and Sastry, P Shanti},
  booktitle=AAAI,
  volume={31},
  number={1},
  year={2017}
}

@inproceedings{ma2020normalized,
  title={Normalized loss functions for deep learning with noisy labels},
  author={Ma, Xingjun and Huang, Hanxun and Wang, Yisen and Romano, Simone and Erfani, Sarah and Bailey, James},
  booktitle=ICML,
  pages={6543--6553},
  year={2020}
}

@article{li2024noisy,
  title={Noisy label processing for classification: A survey},
  author={Li, Mengting and Zhu, Chuang},
  journal={arXiv preprint arXiv:2404.04159},
  year={2024}
}

@inproceedings{choy20194d,
  title={4d spatio-temporal convnets: Minkowski convolutional neural networks},
  author={Choy, Christopher and Gwak, JunYoung and Savarese, Silvio},
  booktitle=CVPR,
  pages={3075--3084},
  year={2019}
}

@article{xiao2022polarmix,
  title={Polarmix: A general data augmentation technique for lidar point clouds},
  author={Xiao, Aoran and Huang, Jiaxing and Guan, Dayan and Cui, Kaiwen and Lu, Shijian and Shao, Ling},
  journal={Advances in Neural Information Processing Systems},
  volume={35},
  pages={11035--11048},
  year={2022}
}

@inproceedings{pan2020poss,
  author={Pan, Yancheng and Gao, Biao and Mei, Jilin and Geng, Sibo and Li, Chengkun and Zhao, Huijing},
  booktitle={2020 IEEE Intelligent Vehicles Symposium (IV)}, 
  title={SemanticPOSS: A Point Cloud Dataset with Large Quantity of Dynamic Instances}, 
  year={2020},
  pages={687-693},
}

@inproceedings{behley2019semantickitti,
  title={SemanticKITTI: A dataset for semantic scene understanding of LiDAR sequences},
  author={Jens Behley and Martin Garbade and Andres Milioto and Jan Quenzel and Sven Behnke and Cyrill Stachniss and Jens Gall},
  booktitle=ICCV,
  year={2019}
}

@inproceedings{wu2024rethinking,
  title={Rethinking attention module design for point cloud analysis},
  author={Wu, Chengzhi and Wang, Kaige and Zhong, Zeyun and Fu, Hao and Zheng, Junwei and Zhang, Jiaming and Pfrommer, Julius and Beyerer, J{\"u}rgen},
  booktitle=ICPR,
  pages={249--267},
  year={2024},
  organization={Springer}
}

@article{LabelImbalance,
  title={WEIGHTED POINT CLOUD AUGMENTATION FOR NEURAL NETWORK TRAINING DATA CLASS-IMBALANCE},
  author={Griffiths, D and Boehm, J},
  journal={The International Archives of Photogrammetry, Remote Sensing and Spatial Information Sciences},
  volume={42},
  pages={981--987},
  year={2019},
  publisher={Copernicus GmbH}
}

@inproceedings{fu2023styleadv,
  title={Styleadv: Meta style adversarial training for cross-domain few-shot learning},
  author={Fu, Yuqian and Xie, Yu and Fu, Yanwei and Jiang, Yu-Gang},
  booktitle={CVPR},
  year={2023}
}

@inproceedings{fu2024cross,
  title={Cross-domain few-shot object detection via enhanced open-set object detector},
  author={Fu, Yuqian and Wang, Yu and Pan, Yixuan and Huai, Lian and Qiu, Xingyu and Shangguan, Zeyu and Liu, Tong and Fu, Yanwei and Van Gool, Luc and Jiang, Xingqun},
  booktitle={ECCV},
  year={2024}
}
\bibliographystyle{IEEEtran}

\end{document}